\documentclass{ieeetj}
\usepackage{cite}
\usepackage{amsmath,amssymb,amsfonts}
\usepackage{graphicx,color}
\usepackage{textcomp}
\usepackage{xcolor}
\usepackage[hidelinks]{hyperref}
\usepackage{cite}
\usepackage{lipsum} 
\usepackage{array}
\usepackage{algpseudocode}
\usepackage{physics}
\usepackage{multirow}
\usepackage{subcaption}
\usepackage{mathtools}
\usepackage{comment}

\def\BibTeX{{\rm B\kern-.05em{\sc i\kern-.025em b}\kern-.08em
    T\kern-.1667em\lower.7ex\hbox{E}\kern-.125emX}}
\AtBeginDocument{\definecolor{tmlcncolor}{cmyk}{0.93,0.59,0.15,0.02}\definecolor{NavyBlue}{RGB}{0,86,125}}
\newcolumntype{M}[1]{>{\centering\arraybackslash}m{#1}}

\usepackage{url}
\usepackage{theorem}
\usepackage{multirow}
\usepackage{pifont}
\usepackage[outline]{contour}
\newcommand{\cmark}{\contour{black}{\textcolor{green}{\ding{51}}}}
\newcommand{\xmark}{\contour{black}{\textcolor{red}{\ding{55}}}}

\def\OJlogo{\vspace{-4pt}$<$Society logo(s) and publication title will appear here.$>$}
\def\seclogo{\vspace{10pt}$<$Society logo(s) and publication title will appear here.$>$}

\def\authorrefmark#1{\ensuremath{^{\textbf{#1}}}}

\begin{document}

\receiveddate{XX Month, XXXX}
\reviseddate{XX Month, XXXX}
\accepteddate{XX Month, XXXX}
\publisheddate{XX Month, XXXX}
\currentdate{XX Month, XXXX}
\doiinfo{XXXX.2022.1234567}

\markboth{Hybrid Layer-Wise ANN-SNN With Surrogate Spike Encoding-Decoding Structure}{Nhan T. Luu {et al.}}

\title{Hybrid ANN-SNN With Layer-Wise Surrogate Spike Encoding-Decoding Structure}

\author{Nhan T. Luu\authorrefmark{1} (Member, IEEE), Duong T. Luu\authorrefmark{2}, Pham Ngoc Nam\authorrefmark{3} (Member, IEEE), Truong Cong Thang\authorrefmark{4} (Senior Member, IEEE)}
\affil{\textit{College of Communication and Information Technology}, \textit{Can Tho University}, Can Tho, Vietnam}
\affil{\textit{Information and Network Management Center}, \textit{Can Tho University}, Can Tho, Vietnam}
\affil{\textit{College of Engineering and Computer Science}, \textit{VinUniversity}, Hanoi, Vietnam}
\affil{\textit{Department of Computer Science and Engineering}, \textit{The University of Aizu}, Aizuwakamatsu, Japan}
\corresp{Corresponding author: Nhan T. Luu (email: luutn@ctu.edu.vn).}

\begin{abstract}
Spiking Neural Networks (SNNs) have gained significant traction in both computational neuroscience and artificial intelligence for their potential in energy-efficient computing. In contrast, artificial neural networks (ANNs) excel at gradient-based optimization and high accuracy. This contrast has consequently led to a growing subfield of hybrid ANN–SNN research. However, existing hybrid approaches often rely on either a strict separation between ANN and SNN components or employ SNN-only encoders followed by ANN classifiers due to the constraints of non-differentiability of spike encoding functions, causing prior hybrid architectures to lack deep layer-wise cooperation during backpropagation. To address this gap, we propose a novel hybrid ANN–SNN framework that integrates layer-wise encode–decode SNN blocks within conventional ANN pipelines. Central to our method is the use of surrogate gradients for a bit-plane-based spike encoding function, enabling end-to-end differentiable training across ANN and SNN layers. This design achieves competitive accuracy with state-of-the-art pure ANN and SNN models while retaining the potential efficiency and temporal representation benefits of spiking computation. To the best of our knowledge, this is the first implementation of a surrogate gradient for bit plane coding specifically and spike encoder interface in general to be utilized in the context of hybrid ANN-SNN, successfully leading to a new class of hybrid models that pave new directions for future research. Source code for our experiments is publicly available at \url{https://github.com/luutn2002/has-8}.
\end{abstract}



\maketitle

\section{INTRODUCTION}
\IEEEPARstart{R}{ecent} decades have witnessed remarkable progress in the domain of artificial intelligence (AI), primarily fueled by the evolution of deep learning techniques \cite{goodfellow2016deep}. Nevertheless, conventional artificial neural networks (ANNs) often struggle to accurately replicate the complex and dynamic characteristics inherent in biological neural systems \cite{nunes2022spiking}. In response to this limitation, spiking neural networks (SNNs) have gained attention as a biologically inspired alternative, exploiting the temporal dynamics of neuronal signaling to better reflect the functioning of the human brain \cite{nunes2022spiking, fang2021deep, luu2024}.

SNNs communicate using discrete spike events, emulating the asynchronous and event-driven nature of biological neural communication \cite{auge2021survey}. The sparsity and temporally precise signaling intrinsic to SNNs make them ideal candidates for designing power-efficient systems capable of real-time processing \cite{rajendran2019low}.

Despite their advantages, SNNs present notable difficulties, particularly in terms of training and optimization. The non-differentiable nature of spike-based activity renders traditional backpropagation techniques unsuitable \cite{nunes2022spiking}. To overcome these obstacles, researchers have proposed alternative training paradigms, including spike-timing-dependent plasticity (STDP) \cite{liu2021sstdp} and surrogate gradient methods \cite{neftci2019surrogate, eshraghian2023training, fang2021deep}, which facilitate the learning process in spiking networks.

Consequentially, several new studies exploring hybrid ANN–SNN models \cite{aydin2024hybrid, kugele2021hybrid, ahmed2024hybrid, negi2024best} have emerged, aiming to combine the strengths of both paradigms to achieve enhanced performance. Although these approaches have demonstrated promising results, they are often implemented as loosely connected modules within a larger architecture and consequently lack deep integration or cooperative interaction across operations. Along side with that, existing gradient-based solutions do not fully resolve the issue of spike signal non-differentiability. Specifically, the encoding function that maps real-valued inputs to discrete spike signals remains non-differentiable. As a result, research directions involving input gradient calculation \cite{czarnecki2017sobolev, ross2018improving} cannot be directly applied from ANN to SNN.

Through our study, we propose a novel approach in which the derivative of spike coding function (specifically bit-plane encoding which had been investigated in previous literatures \cite{luu2024, luu2024improvement}) is replaced with a soft approximation, thereby enabling backpropagation with respect to pre-encoded inputs. This formulation gives rise to a new class of hybrid ANN–SNN models that integrate at the layer level for image processing tasks. Moreover, our empirical evaluation shows that the proposed method achieves competitive and consistent performance compared to state-of-the-art (SOTA) techniques under surrogate gradient training. Contributions of this work are presented as follows:
\begin{itemize}
    \item We introduce a soft approximation method to derive surrogate gradients for spike coding functions in general, and for bit-plane coding in particular. To the best of our knowledge, this represents the first attempt to formulate such approximations for spike coding functions in the context of SNN, opening multiple directions for future research. 
    \item Building upon this derivation, we establish a new class of layer-wise cooperative \underline{h}ybrid \underline{A}NN–\underline{S}NN with \underline{8} bits as timestep models (denoted \textit{HAS-8} in our work) by constructing a completely differentiable encode-decode structure within each utilized SNN block, coupling with its traditional counterpart. Resulted models demonstrate strong performance on image processing tasks in comparison with SOTA baselines.
\end{itemize}

\section{Related works}

\subsection{Spiking neural networks (SNNs)}
SNNs extend traditional ANNs by introducing a biologically inspired mode of computation based on discrete spike events. Unlike ANNs, which rely on continuous activation values and are typically trained using gradient descent methods such as backpropagation \cite{rumelhart1986learning, shrestha2018slayer}, SNNs process information through temporally precise spikes \cite{maass1997networks}. This spike-based, event-driven paradigm enables sparse activity and can significantly reduce energy consumption, especially when implemented on neuromorphic platforms \cite{merolla2014million}. Rather than operating with synchronous, dense real-valued signals like most ANN models, SNNs mimic the behavior of biological neurons, supporting asynchronous and time-resolved communication \cite{gerstner2002spiking}. Their intrinsic ability to handle temporal patterns makes them particularly suitable for sequential data processing and allows the use of biologically motivated learning rules such as STDP \cite{lee2018training}. 

Nonetheless, the non-differentiability of spike events poses a major obstacle to conventional optimization approaches. To address this, training strategies such as surrogate gradient descent \cite{neftci2019surrogate} and ANN-to-SNN conversion methods \cite{rueckauer2017conversion} have been developed to enable more effective learning in SNNs. These innovations highlight both the opportunities and challenges associated with building high-performance, biologically plausible models.

A variety of surrogate gradient formulations have emerged to address the issue of non-differentiability, each offering different balances between biological realism, computational cost, and gradient stability \cite{doi:10.1126/sciadv.adi1480, fang2021deep, zenke2018superspike, esser2015backpropagation, zenke2021remarkable}. These techniques allow for backpropagation-compatible training of spiking layers, enabling the development of deep SNN models capable of handling complex visual recognition tasks \cite{fang2021deep, lee2020enabling, shrestha2018slayer}. Collectively, such advancements have narrowed the performance gap between SNNs and traditional ANNs, while preserving the benefits of sparse, event-driven computation inherent to spiking models.

\subsection{Hybrid ANN-SNN architecture}

In recent years, a significant research thrust has emerged in the domain of SNNs that focuses on creating hybrid architectures which synergistically combine the strengths of traditional ANNs with the efficiency of SNNs \cite{aydin2024hybrid, kugele2021hybrid, ahmed2024hybrid, negi2024best}. The primary motivation behind this approach is to leverage the powerful, gradient-based learning and rich feature extraction capabilities of deep ANNs while capitalizing on the event-driven, low-power computational paradigm of SNNs, attempting to circumvent some of the inherent difficulties in training deep SNNs from scratch.

One of the simplest approaches to hybrid model design is to construct an ensemble consisting of both ANN and SNN components, as demonstrated in \cite{aydin2024hybrid}. This method only requires that the weight gradients in both models be well-defined, thereby enabling joint optimization across the ensemble.  

Another widely adopted strategy involves employing a convolutional SNN as a feature extractor. In this framework, the initial convolutional layers of a well-established SNN architecture are used to process input images asynchronously over multiple timesteps, producing high-level spatiotemporal feature maps. These features are subsequently passed to conventional ANN layers through accumulators, which convert the temporal spike sequences into analog activations. Variants of this approach have been explored through the construction of deep convolutional layers \cite{kugele2021hybrid, negi2024best} as well as the incorporation of spatial attention mechanisms \cite{ahmed2024hybrid}.  

However, the reverse formulation where ANN layers precede SNN components remains challenging due to the non-differentiability of spike generation. In this work, we aim to address this limitation and provide a pathway toward more flexible hybrid architectures, thereby facilitating future research in this direction.

\subsection{Bit plane codes in neural model optimization}
In digital discrete signals, such as images, a bit plane refers to the set of bits that occupy the same positional index within the binary representation of the signal’s numerical values. For example, in an 8-bit representation, the signal can be decomposed into eight distinct bit planes. The first bit plane corresponds to the least significant bits, whereas the eighth bit plane represents the most significant bits. The lowest-order plane captures a coarse yet essential approximation of the signal, while higher-order planes contribute increasingly finer details. Thus, incorporating successive bit planes leads to a progressively refined reconstruction of the original signal.

Bit plane decomposition has found wide application in deep learning, particularly for enhancing efficiency and robustness. In \cite{vorabbi2023input}, bit planes were employed to encode the input layer of Binary Neural Networks (BNNs). The authors demonstrated that bit planes preserve critical spatial information, thereby enabling a reduction in multiply-accumulate (MAC) operations without sacrificing model accuracy or resilience. This result underscores the potential of bit-plane encoding as a means to streamline computation while retaining strong performance.

The role of bit planes in adversarial defense was investigated in \cite{liu2022defending}, where slicing was applied to RGB images. By focusing on the most significant bit planes, the proposed method produced classifiers with improved resistance to adversarial perturbations. This robustness arises from the distinct spatial information captured at different bit levels, which mitigates the effect of adversarial manipulations.

In medical imaging, bit-plane analysis has also proven beneficial. For breast cancer recognition tasks, \cite{chen2019breast} showed that lower-order bit planes are highly noise-sensitive. By excluding these planes before processing with convolutional neural networks (CNNs), the authors achieved superior classification accuracy. This finding highlights the adaptability of bit-plane techniques to domain-specific challenges such as noise suppression, ultimately enhancing model effectiveness in sensitive applications.

While bit plane codes have been extensively studied and applied in ANNs, their utility in SNNs has only recently been investigated. This emerging research focuses on surrogate gradient optimization, where the unique representational properties of these codes are leveraged to facilitate more efficient learning \cite{luu2023blind, luu2024}.

\section{Proposed method}

\subsection{Challenges} \label{challenges}

Deriving surrogate gradients for non-differentiable components of SNN had been a common approach to facilitate gradient-based optimization \cite{fang2021deep, fang2021incorporating, doi:10.1126/sciadv.adi1480}. Similarly, in order to create a hybrid ANN-SNN model with deep layer-wise cooperation, we propose the use of surrogate gradients tailored specifically for the coding function used to convert analog input values into spike trains. 

Designing a surrogate gradient for coding function is a non-trivial problem. Several critical properties must be taken into account, including the determinism, structure, and periodicity of the function. A particularly important consideration is that the coding process should not rely on stochastic sampling. For instance, in rate coding schemes \cite{auge2021survey, heeger2000poisson}, spikes are often generated from a random distribution conditioned on input intensity. While this approach produces statistically meaningful representations, the inherent error \cite{gautrais1998rate} of the sampled spikes introduces uncertainty that makes it difficult to define a stable and differentiable surrogate function for backpropagation.

Instead, we argue that while coding functions don't have to be continuously differentiable, they should ideally be periodic in nature (similarly to Kim et al. \cite{kim2018deep} approach) in order for an approximation to exist. A periodic structure enables the construction of a soft surrogate gradient by approximating the coding behavior using combination of differentiable periodic functions. This soft approximation can then be used in the backward pass, effectively enabling end-to-end training of SNNs while preserving the interpretability and functionality of the original coding scheme.

While the soft approximation derivation of periodical square wave can be considered a correct approach, there are several problems that needed to be addressed before an approximation's gradient can be directly used as surrogate gradient. Considering an example square wave of:
\begin{equation}
f(t) = 
\begin{cases}
1, & \sin(\omega t) \ge 0 \\
-1, & \sin(\omega t) < 0
\end{cases}
=
\operatorname{sgn} \left( \sin(\omega t) \right).
\end{equation}
If we attempt to derive the wave gradient using classical differential rules, we would have:
\begin{equation}
    \frac{df}{dt} = 0, \quad \lim_{t \to t_0^-} f(t) = 1, \quad \lim_{t \to t_0^+} f(t) = -1,
\end{equation}
where the flat segments of the wave (where $f(t)$ is constant at $+1$ or $-1$), the derivative is zero. At the discontinuity points $t = t_0$, the function jumps between $+1$ and $-1$. Since the left-hand and right-hand limits differ, the derivative is undefined at these points. In the sense of generalized functions (distributions), the derivative of the square wave is represented using Dirac delta impulses \cite{jones1968distribution}. For a symmetric $\pm1$ square wave of period $T$, we have:
\begin{equation}
\frac{df}{dt} = 2 \sum_{k=-\infty}^{\infty} (-1)^k \, \delta \!\left( t - \frac{kT}{2} \right), 
\end{equation}
where $\delta(\cdot)$ is the Dirac delta distribution. Each transition contributes an impulse whose integral equals the total jump $\Delta f = 2$. If the square wave is approximated by a smooth transition using a scaled hyperbolic tangent:
\begin{equation}
\begin{aligned}
    & f_\epsilon(t) = \tanh\!\left( \frac{t}{\epsilon} \right),\\
    & \frac{df_\epsilon}{dt} 
= \frac{1}{\epsilon} \, \operatorname{sech}^2\!\left( \frac{t}{\epsilon} \right), \quad \lim_{\epsilon \to 0} \frac{df_\epsilon}{dt} \to \infty,
\end{aligned}
\label{eq:explode}
\end{equation}
where as $\epsilon \to 0$, the slope magnitude near the transition grows without bound, which is consistent with the Dirac delta interpretation. So, if using numerical approximations, the slope can be arbitrarily large, approaching infinity as the transition sharpness increases. Assuming a zero-mean approximation over one period $T$, the variance of the derivative $\frac{df_\epsilon}{dt} $ is defined as \cite{capinski2004measure}:
\begin{equation}
    \begin{aligned}
        \mathrm{Var}\!\left[\frac{df_\epsilon}{dt}\right] &= \frac{1}{T} \int_{-T/2}^{T/2} \left(\frac{df_\epsilon}{dt}\right)^{2} \, dt\\ 
        &= \frac{1}{T} \int_{-T/2}^{T/2} \frac{1}{\epsilon^2} \, \operatorname{sech}^4\!\left(\frac{t}{\epsilon}\right) \, dt\\
    \end{aligned}
\end{equation}
Let $u = \dfrac{t}{\epsilon}$, $dt = \epsilon \, du$, we would have: 
\begin{equation}
    \begin{aligned}
        \mathrm{Var}\!\left[\frac{df_\epsilon}{dt}\right] & = \frac{1}{T \epsilon} \int_{-T/(2\epsilon)}^{T/(2\epsilon)}
\operatorname{sech}^4(u) \, du \\
        & \approx \frac{1}{T \epsilon} \int_{-\infty}^{\infty} \operatorname{sech}^4(u) \, du \quad \text{if} \quad \epsilon \to 0 \Rightarrow \dfrac{T}{2\epsilon} \to \infty\\
        & \approx \frac{4}{3T} \cdot \frac{1}{\epsilon} \quad \text{where} \quad \int_{-\infty}^{\infty} \operatorname{sech}^4(u) \, du = \frac{4}{3}.
    \end{aligned}
\end{equation}
Therefore, as $\epsilon \to 0$, the variance diverges:
\begin{equation}
\lim_{\epsilon \to 0} 
\mathrm{Var}\!\left[\frac{df_\epsilon}{dt}\right] = \infty.
\label{eq:inf_var}
\end{equation}

As implied from Equation \ref{eq:explode} and \ref{eq:inf_var}, it becomes evident that employing soft approximations introduces several unfavorable properties for backpropagation in SNNs:
\begin{itemize}
    \item \textit{Exploding gradients}: when the approximation closely mimics a square wave, the surrogate gradient magnitude diverges toward infinity, leading to severe instability during training.
    \item \textit{High gradient variance}: sharper approximations of the square wave induce significantly larger gradient variance in periodic functions with strong oscillations, which can severely disrupt gradient-based optimization and hinder convergence \cite{gorbunov2020unified, goodfellow2016deep}.
\end{itemize}
To overcome these mentioned challenges, we will propose our solution in following sections.

\subsection{Spike encoding function with surrogate gradient} \label{surrogate_spike}

In our experiments, we focus on bit-plane coding as a candidate coding method. Bit-plane coding \cite{luu2024, luu2024improvement} has the appealing property that each pixel value is encoded into a unique spike pattern, which in our opinion enhances the discriminability of the encoded signals. This encoding approach introduces structured stimulation across neurons in the network similar to traditional deep learning \cite{bengio2013representation} that could lead to a more diverse and informative activations during training. Each bit plane $B_{k, C}$ of $k^{th}$ plane in channel $C$ in coding function \cite{luu2024, luu2024improvement, jahne2005digital} can be formulate as:
\begin{equation}
    B_{k, C}(x, y) = \left\lfloor \frac{I_C(x, y)}{2^k} \right\rfloor \mod 2
\end{equation}
where $I$ is the corresponding pixel intensity. Coded bits on each order $k$ can also be represent as a square wave that toggles at its set of discontinuity points \cite{stromberg2015introduction}, defined as:
\begin{equation}
    \mathcal{D}_k  =  \{\, m\,2^k  :  m\in\mathbb{Z^+} \,\}.
\end{equation}
Hence, over any finite range $[0,L]$ the number of toggles $N_k(L) =  \#\{\,\mathcal{D}_k \cap [0,L]\,\}$ have an upper bound and lower bound of:
\begin{equation}
    \begin{aligned}
        & c_1 \cdot \frac{L}{2^k} \le N_k(L) \le c_2 \cdot \frac{L}{2^k} \quad \forall L > L_0\\ 
        & \text{given} \quad\exists c_1, c_2 > 0, \quad\exists L_0 > 0. \\
    \end{aligned}
\end{equation}
We can also express this using the asymptotically tight upper bound $\Theta$\footnote{Not to be mistaken with Heaviside function $\Theta$ as commonly used in other literature related to SNN.} as:
\begin{equation}
    N_k(L) =  \Theta\!\left(\frac{L}{2^k}\right),\\
\end{equation}
where lower-order bits (small $k$) oscillate more frequently than higher-order bits (large $k$).

Let $f_\epsilon$ be a smooth step (e.g.\ logistic or $\tanh$, similar to approximation method used in Equation \ref{eq:explode}) with derivative $\partial f_\epsilon(t)=\frac{1}{\epsilon}\,\kappa\!\left(\tfrac{t}{\epsilon}\right)$, where $\kappa$ is a fixed, integrable bump function \cite{gehring1976graduate}. A standard soft approximation of $B_{k,C}$ along the intensity axis $i:=I_C(x, y)$ can be written so that its intensity derivative is a superposition of localized bumps at each toggle:
\begin{equation}
\frac{\partial}{\partial i}\,B_{k,C}^{(\epsilon)}(i)\approx\frac{1}{\epsilon}\sum_{m\in\mathbb{Z}} s_m\,\kappa\!\left(\frac{i - m\,2^k}{\epsilon}\right),
\quad s_m\in\{\pm 1\}.
\end{equation}
Consider the period-averaged (or range-averaged) variance of this derivative over $[0,L]$:
\begin{equation}
    \begin{aligned}
    \operatorname{Var}\!\left[\frac{\partial}{\partial i} B_{k,C}^{(\epsilon)}\right]
    &:= \frac{1}{L}\int_{0}^{L}\!\!\left(\frac{\partial}{\partial i} B_{k,C}^{(\epsilon)}(i)\right)^{\!2} \, di \\
    &\approx \frac{1}{L}\,N_k(L)\int_{\mathbb{R}}\!\left(\frac{1}{\epsilon}\,\kappa\!\left(\frac{u}{\epsilon}\right)\right)^{\!2} du \\ &
    = \frac{N_k(L)}{L}\cdot\frac{c_\kappa}{\epsilon}
    = \frac{c_\kappa}{\epsilon\,2^k},
\end{aligned}\end{equation}
where $c_\kappa := \int_{\mathbb{R}} \kappa(u)^2\,du$ depends only on the chosen surrogate. Similarly to the results in Equation \ref{eq:explode} and \ref{eq:inf_var}: 
\begin{itemize}
    \item For fixed smoothing $\epsilon>0$, lower-order bits (small $k$) have larger variance and thus larger surrogate-gradient magnitude; higher-order bits (large $k$) have smaller variance.
    \item For any fixed $k$, the variance scales as $1/\epsilon$ and diverges as $\epsilon\to 0$, recovering the distributional (impulsive) limit.
\end{itemize}

To develop a stable and effective approximation for bit-plane coding without encountering gradient explosion or variance divergence during training, we propose an \textit{order-aware gradient rescaling} strategy. Specifically, we introduce a scaling function applied to the surrogate gradient of bit-plane codes, defined as:
\begin{equation}
    \mathcal{F}\!\left(\partial f_\epsilon(t)\right) 
    = \frac{k}{2^{7-k}} \, \partial f_\epsilon(t),
    \label{eq:rescale_fn}
\end{equation}
where \(k\) denotes the bit-plane index. This scaling function attenuates the excessive variance observed in low-order gradients while preserving sufficient sensitivity for higher-order components. By incorporating this rescaling mechanism, we effectively mitigate the issues highlighted in Equations~\ref{eq:explode} and~\ref{eq:inf_var}, while simultaneously establishing a near-linear correlation between gradient magnitude and the significance of pixel-value representation within the bit string.

From the construction of bit-plane coding \cite{jahne2005digital}, higher-order bits encode larger pixel-value contributions, while lower-order bits primarily capture fine-grained or noise-like variations. By rescaling gradient magnitudes according to the bit-plane index \(k\), we implicitly align gradient updates with pixel-value significance: low-intensity pixels (dominated by low-order bits) contribute minimally to parameter updates, whereas high-intensity pixels (dominated by high-order bits) retain their original gradient scale. The intuition behind this gradient-value alignment is also well supported by analogous practices in conventional deep learning, where input-driven gradient scaling is a common design choice. A prominent example is the ReLU activation function \cite{hahnloser2000digital, goodfellow2016deep}, in which gradients naturally scale with the magnitude of the input. For a simplified case:
\begin{equation}
    \begin{aligned}
        & y = \mathrm{ReLU}(w x), \mathcal{L} = \frac{1}{2} (y - t)^2, \\
        & \frac{\partial \mathcal{L}}{\partial w} =
        \begin{cases}
            (y - t) \cdot x, & \text{if } w x > 0, \\
            0, & \text{otherwise}.
        \end{cases}
    \end{aligned}
\end{equation}
When the neuron is active (\(w x > 0\)), the gradient magnitude scales linearly with \(|x|\), assigning greater influence to inputs with larger activations.

Upon experimenting, we proposed several surrogate gradients for bit-plane coding as follow:
\subsubsection{Sigmoid wrapped sines based surrogate function ($SigSine$)} Bit plane signal can be softly approximate using a sigmoid-wrapped $\sigma$ and a sine function of:
\begin{equation}
    \begin{aligned}
        &\begin{aligned}
            B_{k, C}(x, y) & \approx SigSine(x, y)\\ 
            &= \sigma \left( \alpha\cdot\text{sin}\left(\frac{\pi I_C(x, y)}{2^k}\right)\right),\\
        \end{aligned}\\
        & u(I_C(x, y)) = \alpha\cdot\text{sin}\left(\frac{\pi I_C(x, y)}{2^k}\right), \\ 
        &\nabla_I u = \frac{\alpha \pi}{2^k}\cdot\text{cos}\left(\frac{\pi I_C(x, y)}{2^k}\right),\\
        & \begin{aligned}
        \nabla_I B_{k, C} & \approx \mathcal{F}\left(\nabla_I SigSine\right) \\
        & = \mathcal{F}\left(\sigma(u) \cdot \left( 1- \sigma(u)\right) \cdot \nabla_I u\right) ,\\
        \end{aligned} \\ 
    \end{aligned}
    \label{eq:SigSine}
\end{equation}
where scalar $\alpha \rightarrow -\infty$ would give a sharper transition and a closer approximation to bit plane coding. We also added rescale function $\mathcal{F}(x)$ term to allow gradient rescaling based on input values. In all of our experiment we have $\alpha = -10$. 

\subsubsection{Tanh wrapped sine based surrogate function ($TanhSine$)} Similarly to $SigSine$, we can have another soft approximation by using tanh as a wrap function:
\begin{equation}
    \begin{aligned}
        & \begin{aligned}
        B_{k, C}(x, y) & \approx TanhSine(x, y)  \\
        & = \frac{1}{2}\left(1 + \text{tanh}\left( \alpha\cdot\text{sin}\left(\frac{\pi I_C(x, y)}{2^k}\right)\right)\right),
        \end{aligned}\\
        & u(I_C(x, y)) = \alpha\cdot\text{sin}\left(\frac{\pi I_C(x, y)}{2^k}\right),\\
        &\nabla_I u = \frac{\alpha \pi}{2^k}\cdot\text{cos}\left(\frac{\pi I_C(x, y)}{2^k}\right), \\
        & \nabla_I B_{k, C}  \approx \mathcal{F}\left( \nabla_I TanhSine \right) = \mathcal{F}\left( \text{sech}^2\left(u\right) \cdot \nabla_I u \right), \\
    \end{aligned}
    \label{eq:TanhSine}
\end{equation}
where rescale function $\mathcal{F}(x)$ and scalar $\alpha$ is similarly defined in Equation \ref{eq:SigSine}. We also have $\alpha = -10$.

\subsubsection{Sine based Fourier series ($FourierSine$)} Approximating square waves using Fourier series had been a well known and widely used method \cite{trigub2012fourier}, which can be applied to approximate bit plane coding using a sine based series of:
\begin{equation}
    \begin{aligned}
        & \begin{aligned}
            B_{k, C}(x, y) & \approx FourierSine(x, y)\\
            & = \frac{1}{2} - \sum_{n=1,3,5,...}{\frac{2}{n\pi}\cdot\text{sin}\left(\frac{n \pi I_C(x, y)}{2^k}\right)},\\
        \end{aligned} \\
        & \begin{aligned}
            \nabla_I B_{k, C} & \approx \mathcal{F} \left(\nabla_I FourierSine\right))\\
            & = \mathcal{F} \left(\frac{1}{2} - \sum_{n=1,3,5,...}{\frac{1}{2^{k-1}}\cdot\text{cos}\left(\frac{n \pi I_C(x, y)}{2^k}\right)}\right),\\
        \end{aligned}\\
    \end{aligned}
    \label{eq:FourierSine}
\end{equation}
where rescale function $\mathcal{F}(x)$ is similarly defined in Equation \ref{eq:SigSine}. We have $n = 5$ for all of our experiments.

\subsection{Spike Decoding Functions}

In order for spike signals from SNN to be reused by ANN modules and facilitate deep cross layer cooperation, a decoding function must be involved to process spike to interpretable output. In contrast to spike encoding, differentiable spike decoding functions are already well established and have been widely adopted in the SNN training literature \cite{fang2021deep, fang2021incorporating}. In this section, we introduce the decoding schemes employed in our experiments.

\subsubsection{Rate-based Decoding ($RD$)}

A widely used approach for decoding output spike trains $O[t]$ \cite{fang2021deep, fang2021incorporating} is the rate-based method, which computes the average firing activity across multiple timesteps $t \in T$. Formally, the decoded output $Y$ is obtained as:
\begin{equation}
\begin{aligned}
    & Y = RD(O[t]) = \mathbb{E}_T[O[t]] = \frac{\sum_{t=0}^{T-1}O[t]}{T},\\
    & \frac{\partial Y}{\partial O[t]}=\frac{1}{T} \\
\end{aligned}
\end{equation}

\subsubsection{Bit-plane Decoding ($BPD$)}

In contrast to the non-differentiable encoding process, the decoding of bit-plane representations $O[t]$ into output values $Y$ is fully differentiable. The decoded signal $Y$ is computed as a weighted sum of the bit planes as:
\begin{equation}
    \begin{aligned}
        & Y = BPD(O[t]) = \frac{\sum_{t=0}^{T-1}O[t]\cdot 2^{T-1-t}}{Y_{\text{max}}},\\
        & \frac{\partial Y}{\partial O[t]}=\frac{2^{T-1-t}}{Y_{\text{max}}}, \quad Y_{\text{max}}=255, \quad T = 8\\
    \end{aligned}
\end{equation}
where timestep $T=8$ align with 8 bits in bit planes and $Y_{\text{max}}$ represent max pixel value used for rescaling.

\subsection{HAS-8 model design}

\subsubsection{HAS-8-VGG variants}
\begin{figure*}[t!]
    \centering
    \begin{subfigure}[t]{0.33\textwidth}
        \centering
        \includegraphics[height=0.7\textwidth]{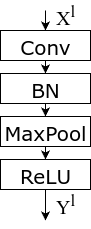}
        \captionsetup{width=0.9\linewidth}
        \caption{Convolutional block of VGG with batch normalization.}
    \end{subfigure}%
    \begin{subfigure}[t]{0.33\textwidth}
        \centering
        \includegraphics[height=0.7\textwidth]{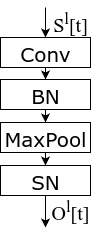}
        \captionsetup{width=0.9\linewidth}
        \caption{Convolutional block of Spiking-VGG with batch normalization.}
    \end{subfigure}%
    \begin{subfigure}[t]{0.33\textwidth}
        \centering
        \includegraphics[height=0.7\textwidth]{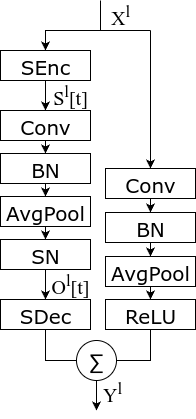}
        \captionsetup{width=0.9\linewidth}
        \caption{Proposed hybrid convolutional block for HAS-8 VGG.}
    \end{subfigure}
    ~
    \begin{subfigure}[t]{0.33\textwidth}
        \centering
        \includegraphics[height=0.5\textwidth]{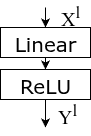}
        \captionsetup{width=0.8\linewidth}
        \caption{MLP block of VGG with ReLU.}
    \end{subfigure}%
    \begin{subfigure}[t]{0.33\textwidth}
        \centering
        \includegraphics[height=0.5\textwidth]{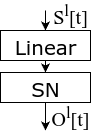}
        \captionsetup{width=0.9\linewidth}
        \caption{MLP block of Spiking-VGG with spiking neuron.}
    \end{subfigure}%
    \begin{subfigure}[t]{0.33\textwidth}
        \centering
        \includegraphics[height=0.5\textwidth]{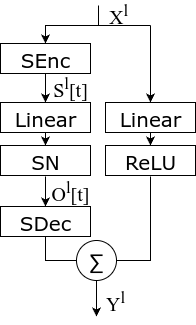}
        \captionsetup{width=0.9\linewidth}
        \caption{Proposed hybrid MLP block for HAS-8 VGG.}
    \end{subfigure}
    \caption{Comparison between VGG implementation of ANN, SNN and our proposed HAS-8-VGG design with spike encoding-decoding and element-wise addition for output fusion.}
    \label{fig:vgg_model_design}
\end{figure*}

Our proposed HAS-8-VGG model is constructed by closely following the design principles of the VGG architecture (both traditional \cite{simonyan2014very} and spiking implementation \cite{sengupta2019going}). In particular, the model consists of multiple deep convolutional layers of comparable depth, which are employed for hierarchical feature extraction. These layers are followed by fully connected multilayer perceptron (MLP) layers that perform the final classification prediction. This design choice ensures that our hybrid model retains the expressive power and strong representational capacity of conventional convolutional architectures.

To enable cross-layer cooperation between ANN and SNN modules, we introduce a mechanism for asynchronous inference across both components. Specifically, the input $X^{l}$ is duplicated and fed in parallel to ANN and SNN processing streams. The SNN branch is composed of Spiking-VGG blocks, which are augmented with differentiable encoding (SEnc) and decoding (SDec) functions to support end-to-end backpropagation. The outputs $O^l[t]$ from the ANN and SNN modules are subsequently fused through an element-wise addition operation, a technique recommended by prior studies \cite{fang2021deep, aydin2024hybrid} to mitigate gradient saturation and stabilize training dynamics. An overview of this architecture is illustrated in Figure~\ref{fig:vgg_model_design}.

In our model design, we adopt the notation \textit{[SEnc-SDec][bx-my-dz]} to represent the configuration of each convolutional encoder. Specifically, the parameter $b$ denotes the base channel count, $m$ denotes the channel multiplier, and $d$ represents the depth index of the block. Accordingly, the number of output channels $C_{out}$ for each convolutional block is computed as:
\begin{equation}
    C_{out} = C_{in} \cdot b \cdot (m^{d}) \quad \text{where} \quad 
    \begin{cases}
        d \in \{1, 2, ..., d_{max}\}, \\
        C_{in} = 3, m=2\\
    \end{cases}
\end{equation}
for majority of HAS-8 model variants. This formulation provides a compact and consistent way to describe the scaling of channel dimensions across different convolutional blocks within the network.

\subsubsection{HAS-8-ResNet variants}

\begin{figure*}[t!]
    \centering
    \begin{subfigure}[t]{0.33\textwidth}
        \centering
        \includegraphics[height=0.8\textwidth]{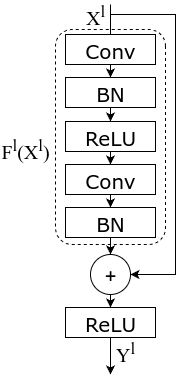}
        \captionsetup{width=0.9\linewidth}
        \caption{Basic block of ResNet.}
    \end{subfigure}%
    \begin{subfigure}[t]{0.33\textwidth}
        \centering
        \includegraphics[height=0.8\textwidth]{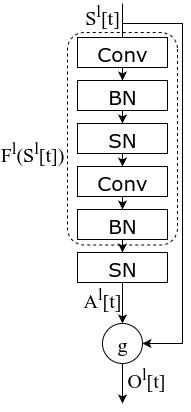}
        \captionsetup{width=0.9\linewidth}
        \caption{Basic block of SEW-ResNet.}
    \end{subfigure}%
    \begin{subfigure}[t]{0.33\textwidth}
        \centering
        \includegraphics[height=0.8\textwidth]{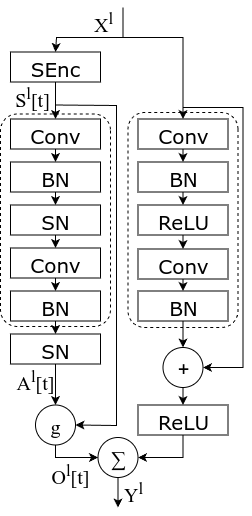}
        \captionsetup{width=0.9\linewidth}
        \caption{Proposed hybrid basic block for HAS-8 ResNet.}
    \end{subfigure}
    ~
    \begin{subfigure}[t]{0.33\textwidth}
        \centering
        \includegraphics[height=0.5\textwidth]{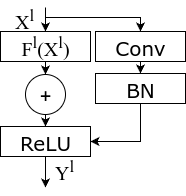}
        \captionsetup{width=0.9\linewidth}
        \caption{Downsample block of ResNet.}
    \end{subfigure}%
    \begin{subfigure}[t]{0.33\textwidth}
        \centering
        \includegraphics[height=0.5\textwidth]{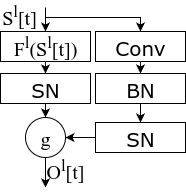}
        \captionsetup{width=0.9\linewidth}
        \caption{Downsample block of SEW-ResNet.}
    \end{subfigure}%
    \begin{subfigure}[t]{0.33\textwidth}
        \centering
        \includegraphics[height=0.5\textwidth]{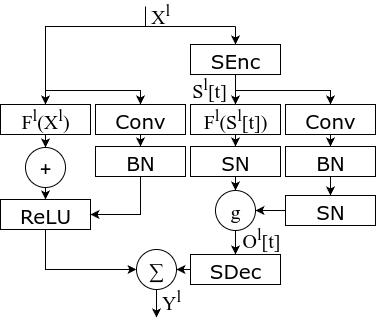}
        \captionsetup{width=0.9\linewidth}
        \caption{Proposed hybrid downsample block for HAS-8 ResNet.}
    \end{subfigure}
    \caption{Comparison between ResNet implementations of ANN, SNN and our proposed HAS-8 ResNet design with spike encoding-decoding and element-wise addition for output fusion.}
    \label{fig:rn_model_design}
\end{figure*}

Following a similar design principle to the VGG-like implementation, we extend our approach to both ResNet18 and SEW-ResNet18 architectures (denoted in Figure~\ref{fig:rn_model_design}). In contrast, we deliberately exclude Spiking-ResNet \cite{hu2021spiking} from our experiments, as previous studies have reported significant gradient instability issues during spike-based backpropagation training\cite{fang2021deep}. For consistency, we also adopt the same architectural notation used in the VGG-based variants to describe our network designs.

\subsection{Overall training settings} 
\label{settings}

To optimize the SNN components of our model, we employed hard-reset IF neurons with arctangent surrogate gradient function with backpropagate through time (BPTT)\cite{fang2021deep, eshraghian2023training} for all benchmarked SNN variants, defined as:
\begin{equation}
    \begin{aligned}
        & \begin{aligned}
            u_i^{(l)}(t) & = \bigl(1 - s_i^{(l)}(t-1)\bigr)\,u_i^{(l)}(t-1)\\
            & + W^{(l)} s^{(l-1)}(t) + b_i^{(l)},\\
        \end{aligned}\\
        &s_i^{(l)}(t) = \Theta\left(u_i^{(l)}(t), V_{\text{th}}\right)\\
        & \Theta(x, V_{th}) =
        \begin{cases}
        1 & \text{if } x - V_{th} \geq 0 \\
        0 & \text{otherwise}
        \end{cases} \\
        & \nabla_x \Theta  = \frac{\alpha}{2\left( 1 + \left( \frac{\pi}{2} \alpha x \right)^2 \right)} \quad \text{where $\alpha=2$}\\
        & \frac{\partial \mathcal{L}}{\partial W^{(l)}} = \sum_{t=1}^{T} \frac{\partial \mathcal{L}}{\partial s^{(l)}(t)} \cdot \frac{\partial s^{(l)}(t)}{\partial u^{(l)}(t)} \cdot \frac{\partial u^{(l)}(t)}{\partial W^{(l)}}\\ 
    \end{aligned}
    \label{heviside}
\end{equation}
where:
\begin{itemize}
    \item $u_i^{(l)}(t)$ is the membrane potential of neuron $i$ in layer $l$ at time $t \in T$,
    \item $s_i^{(l)}(t)$ is the spike output of neuron $i$, similarly with prior layer signal $s_i^{(l-1)}(t)$,
    \item $V_{\text{th}}$ is the threshold voltage,
    \item synaptic weight $W^{(l)}$ and bias $b_i^{(l)}$,
    \item $\Theta$ is the Heaviside function and it's surrogate gradient $\nabla_x \Theta$,
    \item loss function $\mathcal{L}$ and it's gradient with respect to weight $\frac{\partial \mathcal{L}}{\partial W^{(l)}}$.
\end{itemize}
All experiments were conducted using the PyTorch framework~\cite{paszke2019pytorch}, along with the SpikingJelly library \cite{doi:10.1126/sciadv.adi1480}.

Inspired by traditional deep learning approach in model fine-tunning \cite{luu2023blind, luu2024universal} and prior work in surrogate SNN optimization \cite{fang2021deep, luu2024improvement}, optimization for all ANN and SNN variants was carried out using the Adam optimizer \cite{kingma2014adam}, with a fixed learning rate of \( lr = 10^{-3} \), decay rate of \( \lambda = 10^{-3} \) and momentum parameters \( \beta = (0.9, 0.999) \), consistently applied across all experiments. For the loss function, we used the Cross-Entropy Loss~\cite{paszke2017automatic}, expressed as:
\begin{equation}
    \begin{aligned}
        \theta^\ast &= \arg\min_\theta \sum_{n=1}^N \sum_{k=1}^K -y_{nk} \log f(x_n ; \theta)_k \\
        &= \arg\min_\theta \mathcal{L}_{CE}(y, \hat{y})
    \end{aligned}
    \label{eq:cel}
\end{equation}

To ensure reproducibility and consistency with prior researches, we followed the original dataset partitioning wherever available. For datasets lacking predefined splits, we adopted an 80/20 training-validation split. Small-scale image datasets resolution is kept as is and ones with high pixel counts (exceed 224 pixels per resolution dimension) are resized to 224x224. All experiments were conducted using fixed random seeds~\cite{picard2021torch} for reproducibility, and training was carried out over 100 epochs on an NVIDIA RTX 3090 GPU with 24 GB of VRAM.

\section{Experimental results}

\begin{table*}[t!]
    \caption{Ablation study of HAS-8-VGG[b16-m2-d4] and HAS-8-ResNet[b32-m2-d4] with and without gradient rescaling on CIFAR-10. Inclusion of rescaling term tend to perform better in practice. We also compare our performance with some other SNN/ANN models within the same training configuration. Best performance per variant is denoted in bold.}
    \centering
    \resizebox{0.9\textwidth}{!}{
        \begin{tabular}{|M{2.71cm}|c|c|c|c|c|c|c|c|c|}
        \hline
        \multirow{3}{*}{\bf Model variants} & \multirow{3}{*}{\bf Timestep $T$} & \multirow{3}{*}{\bf \shortstack{Parameter \\ count}} & \multirow{3}{*}{\bf \shortstack{Gradient \\ scaling}} & \multicolumn{3}{c|}{\bf $\nabla_x$ SEnc} & \multicolumn{2}{c|}{\bf SDec} & \multirow{3}{*}{\bf Accuracy}\\ 
        \cline{5-9}
         & & & & \shortstack{$Sig$\\$Sine$} & \shortstack{$Tanh$\\$Sine$} & \shortstack{$Fourier$\\$Sine$} & $RD$ & $BPD$ & \\
         \hline
         SEW-ResNet18 (RC) & \multirow{3}{*}{10} & \multirow{4}{*}{11.7M} & \multirow{4}{*}{\xmark} & \multirow{4}{*}{\xmark} & \multirow{4}{*}{\xmark} & \multirow{4}{*}{\xmark} & \multirow{3}{*}{\cmark} & \multirow{4}{*}{\xmark} & 70.69 \\
         \cline{1-1}\cline{10-10}
         SEW-ResNet18 (TTFS) & & & & & & & &  & 72.24 \\
         \cline{1-1}\cline{10-10}
         SEW-ResNet18 (PC)& & & & & & & & & 74.60 \\
         \cline{1-2}\cline{8-8}\cline{10-10}
        ResNet18 & \xmark & & &  &  & & \xmark & & 75.89 \\
        \hline
        \shortstack{\multirow{10}{*}{\bf HAS-8-VGG} \\ \multirow{12}{*}{\bf [b16-m2-d4]}\\ \multirow{14}{*}{\bf (Ours)}} & \multirow{12}{*}{8} & \multirow{12}{*}{9.69M} & \multirow{6}{*}{\xmark} & \cmark & \xmark & \xmark & \multirow{3}{*}{\cmark} & \multirow{3}{*}{\xmark} & 65.53 \\
        \cline {5-7}\cline {10-10}
        & & & & \xmark & \cmark & \xmark & & & 64.01 \\
        \cline {5-7}\cline {10-10}
        & & & & \xmark & \xmark & \cmark & & & 66.12 \\
        \cline {5-10} 
        & & & & \cmark & \xmark & \xmark & \multirow{3}{*}{\xmark} & \multirow{3}{*}{\cmark} & 70.27 \\
        \cline {5-7}\cline {10-10}
        & & & & \xmark & \cmark & \xmark & & & 70.05 \\
        \cline {5-7}\cline {10-10}
        & & & & \xmark & \xmark & \cmark & & & 71.93 \\
        \cline {4-10} 
        & & & \multirow{6}{*}{\cmark} & \cmark & \xmark & \xmark & \multirow{3}{*}{\cmark} & \multirow{3}{*}{\xmark} & 80.44 \\
        \cline {5-7}\cline {10-10} 
        & & & & \xmark & \cmark & \xmark & & & 80.03 \\
        \cline {5-7}\cline {10-10} 
        & & & & \xmark & \xmark & \cmark & & & 81.30 \\
        \cline {5-10} 
        & & & & \cmark & \xmark & \xmark & \multirow{3}{*}{\xmark} & \multirow{3}{*}{\cmark} & 81.15 \\
        \cline {5-7}\cline {10-10} 
        & & & & \xmark & \cmark & \xmark & & & 81.52 \\
        \cline {5-7}\cline {10-10} 
        & & & & \xmark & \xmark & \cmark & & & \textbf{81.58}\\
        \hline 
        \shortstack{\multirow{10}{*}{\bf HAS-8-ResNet} \\ \multirow{12}{*}{\bf [b32-m2-d4]}\\ \multirow{14}{*}{\bf (Ours)}} & \multirow{12}{*}{8} & \multirow{12}{*}{5.63M} & \multirow{6}{*}{\xmark} & \cmark & \xmark & \xmark & \multirow{3}{*}{\cmark} & \multirow{3}{*}{\xmark} & 17.61 \\
        \cline {5-7}\cline {10-10}
        & & & & \xmark & \cmark & \xmark & & & 18.22 \\
        \cline {5-7}\cline {10-10}
        & & & & \xmark & \xmark & \cmark & & & 20.09 \\
        \cline {5-10}
        & & & & \cmark & \xmark & \xmark & \multirow{3}{*}{\xmark}& \multirow{3}{*}{\cmark}& 24.22 \\
        \cline {5-7}\cline {10-10}
        & & & & \xmark & \cmark & \xmark & & & 25.31 \\
        \cline {5-7}\cline {10-10}
        & & & & \xmark & \xmark & \cmark & & & 24.57 \\
        \cline {4-10}
        & & & \multirow{6}{*}{\cmark} & \cmark & \xmark & \xmark & \multirow{3}{*}{\cmark} & \multirow{3}{*}{\xmark}& 59.90 \\
        \cline {5-7}\cline {10-10}
        & & & & \xmark & \cmark & \xmark & & & 48.38 \\
        \cline {5-7}\cline {10-10}
        & & & & \xmark & \xmark & \cmark & & & 73.35 \\
        \cline {5-10}
        & & & & \cmark & \xmark & \xmark & \multirow{3}{*}{\xmark} & \multirow{3}{*}{\cmark} & 61.39 \\
        \cline {5-7}\cline {10-10}
        & & & & \xmark & \cmark & \xmark & & & 63.23 \\
        \cline {5-7}\cline {10-10}
        & & & & \xmark & \xmark & \cmark & & & \textbf{79.20}\\
        \hline
        \end{tabular}
    }
    \label{tab:ablation}
\end{table*}

\subsection{Preliminary studies}

In this section, we present preliminary studies on 2 of our proposed architectures: HAS-8-VGG[b16-m2-d4] and HAS-8-ResNet[b32-m2-d4] (which is the smallest working variants upon empirical evaluation in comparison with the well-known ResNet18) under different training settings and configurations on small scale CIFAR-10\cite{alex2009learning} dataset. Specifically, we investigate the impact of various surrogate gradient spike coding functions on model performance. To provide a comprehensive comparison, we also include several well-known ANN and SNN baselines. These include ResNet18 \cite{he2016deep} (standard ANN implementation) and SEW-ResNet18 \cite{fang2021deep}, an SNN variant that supports multiple spike coding schemes such as rate coding (RC) \cite{heeger2000poisson}, time-to-first-spike (TTFS) \cite{auge2021survey}, and weighted phase coding (PC) \cite{kim2018deep}, we aim to highlight the performance contrast between traditional ANN models, pure SNN models, and our proposed hybrid HAS-8 architecture.

As shown in Table~\ref{tab:ablation}, there is a clear and significant performance gap between hybrid models that utilize the gradient scaling function $\mathcal{F}(x)$ within surrogate gradient implementations and those that do not. The largest recorded performance improvement attributed to $\mathcal{F}(x)$ is observed in VGG-based variants, with a gain of up to $17.57\%$, while an even more substantial improvement of $61.59\%$ is reported for ResNet-based variants. 

Furthermore, our HAS-8 variants consistently achieve higher performance while maintaining significantly lower parameter counts compared to conventional architectures. Specifically, the HAS-8-VGG variant only requires approximately 2M trainable parameters less when compared to ANN and SNN ResNet, whereas in the HAS-8-ResNet variant this gap improve to around 6M parameters. Despite this considerable reduction in model size, our proposed models outperform larger baselines, demonstrating the efficiency of the HAS-8 architecture. This result encourage us to further experiment with this variants.

\begin{table*}[t!]
    \caption{Cross-dataset evaluation of HAS-8-VGG[$FourierSine$-$BPD$] and HAS-8-ResNet[$FourierSine$-$BPD$] variants with other well known SOTA models, both ANN and SNN implementations. Best performance per variant is denoted in bold, second best is underlined and N/A entries are where result not available. Note that ImageNet require a more specific training setting.}
    \centering
    \resizebox{0.9\textwidth}{!}{
        \begin{tabular}{|M{2.5cm}|c|c|c|c|c|c|c|}
        \hline
        \multirow{2}{*}{\bf Model variants} & \multirow{2}{*}{\bf \shortstack{Timestep \\ $T$}} & \multirow{2}{*}{\bf \shortstack{Parameter \\ count}} & \multicolumn{5}{c|}{\bf Top 1 accuracy per dataset (\%)}\\ 
        \cline{4-8}
        & & & CIFAR-10 & CIFAR-100 & Caltech-101 & Caltech-256 & ImageNet\\
        \hline
        SEW-ResNet18 (PC) & \multirow{2}{*}{10} & 11.7M & 74.60 & 42.61 & 64.21 & 40.43 & 51.47\\
        \cline{1-1}\cline{3-8}
        SEW-ResNet50 (PC) & & 23.6M & 73.96 & 41.52 & 64.32 & 34.56 & N/A\\
        \hline
        ResNet18 & \multirow{2}{*}{\xmark} & 11.7M & 75.89 & \textbf{51.11} & \bf 71.24 & 42.35 & \bf 52.43\\
        \cline{1-1}\cline{3-8}
        ResNet50 & & 23.6M & 76.49 & 43.10 & \underline{70.61} & 37.89 & N/A\\
        \hline
        \bf HAS-8-VGG [b16-m2-d4] (Ours) & \multirow{7}{*}{8} & 9.69M & \bf 81.58 & 45.00 & 57.75 & 34.83 & 40.79\\
        \cline{1-1} \cline{3-8}
        \bf HAS-8-VGG [b24-m2-d4] (Ours) & & 19.3M & 77.19 & 43.78 & 61.38 & 32.51 & N/A\\
        \cline{1-1} \cline{3-8}
        \bf HAS-8-ResNet [b32-m2-d4] (Ours) & & 5.63M & 79.20 & 46.99 & 68.70 & \underline{42.90} & \underline{52.30}\\
        \cline{1-1} \cline{3-8}
        \bf HAS-8-ResNet [b64-m2-d4] (Ours) & & 22.4M & \underline{79.90} & \underline{48.23} & 67.95 & \textbf{43.11} & N/A\\
        \hline
        \end{tabular}
    }
    \label{tab:cross-dataset-eval}
\end{table*}

\subsection{Cross-datasets evaluation}

As delineated in Table II, we conducted a comprehensive cross-dataset evaluation of our proposed HAS-8-VGG and HAS-8-ResNet models against several SOTA baselines, including standard ANNs such as ResNet18/50 and SNN implementations such as SEW-ResNet18/50. Our benchmarking datasets consist of CIFAR-10\cite{alex2009learning}, CIFAR-100\cite{alex2009learning}, Caltech-101\cite{li_andreeto_ranzato_perona_2022}, Caltech-256\cite{griffin_holub_perona_2022}, and ImageNet\cite{deng2009imagenet}.

On the CIFAR-10 dataset, our HAS-8-VGG[b16-m2-d4] model achieves SOTA performance, securing a top-1 accuracy of 81.58\%. This result is particularly noteworthy as it surpasses not only the SNN baselines but also the standard ANN ResNet variants. For instance, our model outperforms the best-performing baseline, ResNet18 (75.89\%), by a significant margin of 5.69\% while utilizing fewer parameters (9.69M vs. 11.4M). While the standard ResNet18 ANN demonstrates the highest accuracy on CIFAR-100 (51.11\%) and Caltech-101 (71.24\%), our models remain highly competitive. Our HAS-8-ResNet [b64-m2-d4] variant achieves the second-best result on CIFAR-100 with 48.23\% accuracy. More importantly, on the challenging Caltech-256 dataset, our proposed models claim the top two positions. The HAS-8-ResNet[b64-m2-d4] and HAS-8-ResNet[b32-m2-d4] variants achieve the best and second-best accuracies of 43.11\% and 42.90\%, respectively.

While optimized variants are able to converge on smaller datasets under the settings described in Section~\ref{settings}, training on ImageNet presents additional challenges. Prior studies have shown that Adam-based optimizers struggle to converge effectively on very large-scale classification tasks, where stochastic gradient descent (SGD) typically achieves superior results\cite{zhou2020towards}. Furthermore, without a more sophisticated weight decay scheduling scheme, extended training can experience instability, which further complicates fine-tuning on datasets as large as ImageNet. To address these issues, we adopt a more sophisticated training scheme for all benchmarked models on ImageNet, employing SGDR with an initial learning rate of $lr = 0.1 \times (\text{batch size}/256)$ (batch size similarly defined in Section~\ref{settings})\footnote{In large‐scale training experiments, it is common to use SGDR and initialize the learning rate around 0.1 when using a batch size of about 256\cite{goyal2017accurate}. When batch size is lower, smaller initial learning rates are typically necessary to ensure stable convergence and avoid divergence or suboptimal solutions\cite{masters2018revisiting}.}, combined with Cosine Annealing scheduling method and warm restart from $lr/10$ \cite{loshchilov2016sgdr}, along with Nesterov momentum set to $nesterov=0.9$. 

Upon validation with ImageNet, our HAS-8-ResNet[b32-m2-d4] model achieves the second highest accuracy among all evaluated variants, reaching 52.30\% and slightly falls behind ResNet18 with 52.43\%. This surpasses the standard SEW-ResNet18 baseline of 51.47\% despite using less than half of the parameters (5.63M vs. 11.7M). Notably, even the compact HAS-8-VGG[b16-m2-d4] model maintains competitive performance at 49.70\%, again highlighting the parameter efficiency of our approach\footnote{Total training time of variants that involve SNN modules (SEW-ResNet and HAS-8) on ImageNet is over 1 month per model and pure ANN (ResNet) is around 15 days, given our hardware setting mentioned in Section~\ref{settings}.}.

A another key finding is the consistent and significant performance gap between our HAS models and the SEW-ResNet SNN baselines. Across all five datasets, our models outperform their SEW counterparts, often by a large margin (e.g., 81.58\% vs. 74.60\% on CIFAR-10, 52.30\% vs. 51.47\% on ImageNet). Crucially, our models achieve this with a smaller number of timesteps ($T=8$ vs. $T=10$ for SEW), indicating a lower inference latency and greater computational efficiency, which is a critical advantage for SNNs. The compact HAS-8-ResNet[b32-m2-d4] model, with only 5.63M parameters, provides a compelling trade-off between model size and performance, outclassing all SEW models across the board.

\section{Computational efficiency}

\begin{table}[htb!]
    \centering
    \caption{Comparing total MACs operation of our method with existing SNN and ANN architectures over an input image of size 224x224. For SNN modules only MACs operation over 1 timestep is considered.}
    \resizebox{\linewidth}{!}{
    \begin{tabular}{|M{2.8cm}|*{4}{M{1.3cm}|}} 
    \hline
    \bf Model & \bf Model type & \bf Parameter count & \bf MACs\\
    \hline
    SEW-ResNet18\cite{fang2021deep} & SNN & 11.7M & \underline{1.81G}\\
    \hline
    SEW-ResNet50\cite{fang2021deep}  & SNN & 25.6M & 4.09G\\
    \hline
    ResNet18 & ANN & 11.7M & \underline{1.81G}\\
    \hline
    ResNet50 & ANN & 25.6M & 4.09G\\
    \hline
    \bf HAS-8-VGG [b16-m2-d4] (Ours) & Hybrid & 9.79M & 4.34G\\
    \hline
    \bf HAS-8-VGG [b24-m2-d4] (Ours) & Hybrid & 19.3M & 9.62G\\
    \hline
    \bf HAS-8-ResNet [b32-m2-d4] (Ours) & Hybrid & 5.63M & \textbf{1.16G}\\
    \hline
    \bf HAS-8-ResNet [b64-m2-d4] (Ours) & Hybrid & 22.4M & 3.64G\\
    \hline
    \end{tabular}}
    \label{tab:macs}
\end{table}

In addition to performance benchmarking, we also evaluate the computational efficiency of our proposed HAS-8 models in comparison to conventional ANN and SNN architectures. A common metric for assessing efficiency is the number of multiply-accumulate (MAC) operations, which directly correlates with both energy consumption and inference latency. For SNNs, we report the MACs corresponding to a single timestep, as is standard in prior work. Table~\ref{tab:macs} summarizes the total MACs of all evaluated models for an input image of size $224\times 224$.  

From the results, we observe that our HAS-8 models achieve a favorable balance between accuracy and computational cost. Notably, the HAS-8-ResNet[b32-m2-d4] variant requires only 1.16G MACs, representing the lowest computational overhead among all compared models. This efficiency is particularly striking when contrasted with the ANN ResNet50 baseline, which requires 4.09G MACs and is over 3.5$\times$ higher despite using a similar number of parameters. Even the compact ResNet18 baseline demands 1.81G MACs, which is still significantly greater than our HAS-8-ResNet[b32-m2-d4].  

Our larger HAS-8-ResNet[b64-m2-d4] model further demonstrates the scalability of our approach. With 3.64G MACs, it achieves computational efficiency comparable to ResNet50 (4.09G) but with improved flexibility and hybrid SNN-ANN characteristics. Similarly, the HAS-8-VGG variants, while requiring higher MACs (4.34G and 9.62G, respectively), still remain competitive with their ANN counterparts given the relative improvements in accuracy reported in Table~\ref{tab:cross-dataset-eval}.  

An important observation is that traditional SNN baselines, such as SEW-ResNet18 and SEW-ResNet50, show relatively higher MACs compared to our HAS models when accounting for equivalent performance. For instance, SEW-ResNet50 consumes 4.09G MACs per timestep, yet fails to match the accuracy levels of our HAS-8-ResNet[b32-m2-d4], which is not only more accurate but also more energy-efficient.  

Overall, these findings highlight that our HAS framework is capable of significantly reducing computational cost without sacrificing performance. The ability of HAS-8-ResNet[b32-m2-d4] to achieve the lowest MAC count establishes it as a highly energy-efficient alternative to both ANN and SNN baselines, underscoring the suitability of our approach for deployment in resource-constrained environments.

\section{Conclusion}

In this work, we introduced a novel bit-plane coding function integrated with a surrogate gradient-based spike encoder, enabling fully differentiable training for deeply coupled hybrid ANN-SNN architectures. By formulating a differentiable interface between spike encoding and decoding processes, our approach effectively overcomes the long-standing challenge of gradient discontinuity in spike-based backpropagation. Leveraging a strict 8-timestep coding strategy, we demonstrate that our method facilitates seamless end-to-end optimization while maintaining compatibility with standard gradient-based learning techniques. Our findings indicate that such differentiable spike encoding not only narrows the performance gap between SNNs and ANNs on computer vision tasks but also opens new research directions toward the development of spiking neural architectures.

\bibliographystyle{IEEEtran}
\bibliography{refs}

\vfill\pagebreak

\end{document}